\documentclass{article}
\usepackage{spconf,amsmath,graphicx}
\usepackage[normalem]{ulem}
\useunder{\uline}{\ul}{}
\usepackage{hyperref}       
\usepackage{url}            
\usepackage{amsmath}
\usepackage{wrapfig}
\usepackage{subcaption}
\usepackage{multirow}
\usepackage{booktabs}       
\usepackage{amsfonts}       
\usepackage{nicefrac}       
\usepackage{microtype}      
\usepackage{graphicx}
\usepackage{amssymb}
\usepackage{enumitem}
\usepackage{mathtools}
\usepackage{textcomp}
\usepackage{multirow}
\usepackage{adjustbox}
\usepackage[misc]{ifsym}
\usepackage{comment}
\makeatletter
\newcommand\footnoteref[1]{\protected@xdef\@thefnmark{\ref{#1}}\@footnotemark}
\makeatother
\usepackage{amssymb}
\usepackage{amsmath}
\usepackage{booktabs}

\DeclareMathOperator*{\argmin}{arg\,min}

\DeclareMathOperator*{\len}{len}

\newcommand{\movingaverage}{\model{MA}}

\newcommand{\model}[1]{\texttt{#1$\,$}}

\def\drawplusplus#1#2#3{\hbox to 0pt{\hbox to #1{\hfill\vrule height #3 depth
      0pt width #2\hfill\vrule height #3 depth 0pt width #2\hfill
      }}\vbox to #3{\vfill\hrule height #2 depth 0pt width
      #1 \vfill}}

\title{Auxiliary Quantile Forecasting with Linear Networks}

\twoauthors
  {Shayan Jawed}
        {Machine Learning Lab \\ University of Hildesheim \\ Germany \\ shayan@ismll.uni-hildesheim.de}
  {Lars Schmidt-Thieme}
        {Machine Learning Lab \\ University of Hildesheim \\ Germany \\ 
        schmidt-thieme@ismll.uni-hildesheim.de}

\begin{document}
%
\maketitle

\begin{abstract}
We propose a novel multi-task method for quantile forecasting with shared Linear layers. Our method is based on the Implicit quantile learning approach, where samples from the Uniform distribution $\mathcal{U}(0, 1)$ are reparameterized to quantile values of the target distribution. We combine the implicit quantile and input time series representations to directly forecast multiple quantile estimations for multiple horizons jointly. Prior works have adopted a Linear layer for the direct estimation of all forecasting horizons in a multi-task learning setup. We show that following similar intuition from multi-task learning to exploit correlations among forecast horizons, we can model multiple quantile estimates as auxiliary tasks for each of the forecast horizon to improve forecast accuracy across the quantile estimates compared to modeling only a single quantile estimate. We show learning auxiliary quantile tasks leads to state-of-the-art performance on deterministic forecasting benchmarks concerning the main-task of forecasting the 50$^{th}$ percentile estimate.
\end{abstract}
\begin{keywords}
Implicit Quantile Networks, Multi-task Learning, Linear Networks
\end{keywords}
\section{Introduction}
\label{sec:intro}
Recent methods model the long-term forecasting problem as a multi-task learning problem where a Linear layer directly forecasts for multiple forecast horizons from a shared encoded representation \cite{zeng2022transformers,wu2021autoformer,zhou2021informer}. Remarkably, it was shown in \cite{zeng2022transformers} that even simple linear encoded representations combined with such a Linear layer can achieve state-of-the-art performance.

It is well-known that Multi-task learning can exploit shared latent representations among tasks improving performance across all tasks, rather learning tasks in isolation. 

Following similar intuition of learning rich latent correlation representations between tasks, we propose learning multiple auxiliary tasks of quantile estimations for multiple forecast horizons. For example, exploiting the correlations among the learning tasks of modeling the 10$^{th}$, 50$^{th}$ and the 90$^{th}$ percentile estimations for each of the multiple forecast horizons. 

Quantile forecasting is a well-established probabilistic forecasting method for robustly modeling probabilistic outputs \cite{koenker2001quantile}. We can broadly categorize the prior work among two major categories, firstly we note, Multi-Quantile networks (MQ-RNN) \cite{wen2017multi,park2021learning} that learn to output a discrete set of multiple quantile estimations per forecasting horizon jointly. These methods utilize a base RNN Encoding module and shared Linear layers for quantile forecasting across multiple horizons. Notably, the limitation with this approach is that the set of quantile levels to model remains fixed throughout training. Implicit Quantile Networks (IQNs) solve for this limitation by learning the full quantile function of the data distribution \cite{dabney2018implicit, gouttes2021probabilistic}. The intuition is to embed a randomly sampled quantile level from the unit uniform distribution within the network and learn to output the quantile estimation corresponding to this level. The network learns by minimizing the Quantile loss, which is also parameterized with the same quantile level sampled. Therefore, by successively sampling random quantile levels in the course of optimization, the network converges to modeling for the quantile levels across the unit uniform distribution. 

In our paper, we combine the ideas from these two well-established research streams, and propose to implicitly model multiple quantile estimations per forecast horizon with shared Linear layers. In each gradient update, we sample multiple quantile levels from $\mathcal{U}(0, 1)$ for each forecast horizon. This leads to learning auxiliary quantile estimations tasks, and exploiting latent correlations among these leads to more accurate modeling of the main task of modeling the 50$^{th}$ percentile estimation. This differentiates our work from IQN based learning, where only a single quantile estimate is learned per forecast horizon per gradient update. Our contributions are:

\begin{itemize}
    \item We propose a novel multi-task learning method, that learns auxiliary quantile estimation tasks per forecast horizon with the well-established Implicit quantile learning framework and shared Linear layers.
    \item We propose parameter sharing for the forecasting Linear layer that exploits a shared encoded time series representation combined with implicit quantile representations to forecast multiple quantile estimations without significant increase in the number of learnable parameters.
    \item We propose auxiliary quantile augmentation learning tasks for the given input. Our model reconstructs the input time series as an auxiliary-task and estimates additional quantiles for the input with shared Linear layers as further auxiliary tasks.
    \item We tune the number of additional auxiliary quantile forecasting and reconstruction tasks as a hyperparameter, and model up-to thousand learning tasks jointly that leads to superior performance compared to deterministically only learning the least squares estimates. 
    \item We perform extensive experiments to validate the proposed model's performance compared to several state-of-the-art forecasting baselines on established benchmarks.
\end{itemize}

\section{Methodology}
We consider $N$ related multivariate time series data $\mathbf{X} \in \mathbb{R}^{T \times N \times C}$ where each time series $X^n \in \mathbb{R}^{T \times C}$ is noted for a total of $t=[1,..\tau,...,T]$ timesteps and $\tau$ partitions the observations in the input range, $X_{1:\tau}$ and the multiple horizon indexes for forecasting, $Y_{\tau:T}$. $C$ denotes the number of channels observed for the time series. We assume all channels have been observed for the same timesteps with regular frequency.
\subsection{Implicit Quantile Embedding}
Following the learning framework proposed in \cite{dabney2018implicit}, we embed quantile levels within the forecasting network such that it can forecast exactly the same quantile estimations for the forecast horizons. The implicit quantile embedding module embeds the $i^{th}$ uniformly at random sampled quantile level, scalar, $\alpha_i \in \mathcal{U}(0, 1)$ with two learnable parameters: a weight parameter and an additive bias. The transformation: 
\begin{equation}
    \alpha_i' = \alpha_i W_{iqn} + b_{iqn} 
\end{equation}
Where $W_{iqn} \in \mathbb{R}$ represents the learnable parameter and $b_{iqn} \in \mathbb{R}$ the bias. Given our objective is to forecast multiple $M$ many quantile levels, $\alpha_{1:M} \in \mathbb{R}^M$, we utilize a shallow embedding per quantile level. This allows a scalable embedding for many quantile levels and minimizes the learnable parameters for the multiple quantile levels that are to be embedded. Notably, we do not use a non-linear activation either. 
\subsection{Time Series Input Preprocessing}
Given the quantile embedding, we sum it with the raw time series and then decompose the resulting summation following previous work \cite{zeng2022transformers}.
\subsubsection{QDLinear} 
The time series can be decomposed into a trend component by a moving average kernel. Differencing the trend component from the input leads to the remainder seasonal component. We can express the operation mathematically as follows: 
\begin{equation}
\begin{aligned}
    Z_i &= X_{1:\tau} + \alpha'_i \\ 
    Z_{i,trend} &= \movingaverage(Z_i, w) \\ 
    Z_{i,season} &= Z_i - Z_{i,trend}
\end{aligned}
\label{eq:qdlinear}
\end{equation}
Where, $Z_i$ denotes the quantile embedding added to the input and the function \movingaverage$(.,w)$ calculates the moving average with a specified kernel length $w$ as hyperparameter. Note that this preprocessing is similar to prior work \cite{zeng2022transformers,wu2021autoformer,zhou2022fedformer}, with the notable difference of adding the implicit quantile embedding to the raw time series input firstly. Given that we sample multiple quantile levels, $\alpha_{1:M}$, for each quantile level embedded, we repeat the operations in block Eq. \ref{eq:qdlinear}, each with a separate copy of raw time series input, leading to different decomposed $\{Z_{i,trend},Z_{i,season}\}_{i=1:M}$ components, each representing a decomposition with a different quantile embedding $\alpha'_i$.
\begin{table*}[h]
\vspace{-0.0cm}
\begin{center}
\scalebox{0.8}
{
\begin{tabular}{|c|c|c|c|c|c|c|c|c|c|} \hline
Datasets      & ETTh1 & ETTh2       & ETTm1  & ETTm2        & Traffic     & Electricity  & Exchange-Rate & Weather      & ILI      \\ \hline
Channels      & 7 & 7 & 7         & 7            & 862         & 321          & 8             & 21          & 7      \\ \hline
Observations     & 17,420 & 17,420     & 69,680   & 69,680     & 17,544      & 26,304       & 7,588         & 52,696       & 966    \\
\hline
\end{tabular}
}
\end{center}
\vspace{-0.4cm}
\caption{Multivariate time series dataset statistics \cite{zeng2022transformers,zhou2021informer}}
\label{tab:datasets}
\end{table*}
\subsubsection{QNLinear} 
In case of a distribution shift in the time series data, another preprocessing was proposed in \cite{zeng2022transformers}. The idea is to normalize the time series input through differencing the last time series observation from the entire series before embedding it in the network. Similar to before, we first add the quantile representation to the time series input before the preprocessing:
\begin{equation}
\begin{aligned}
    Z_i &= X_{1:\tau} + \alpha'_i \\ 
    Z_{i,norm} &= Z_i - X_{\tau}
\end{aligned}
\label{eq:qnlinear}
\end{equation}
For QNLinear, we also create multiple copies $\{Z_{i,norm}\}_{i=1:M}$, each representing a normalized copy of the time series input with a different quantile embedding $\alpha'_i$. 
\subsubsection{QLinear} 
Following previous work \cite{zeng2022transformers}, we also design the quantile prediction network with a base one-layer neural network that takes in a quantile embedded summed time series input, as expressed via first Eq. from block Eq. \ref{eq:qdlinear}, \ref{eq:qnlinear}, deriving $\{Z_{i,linear}\}_{i=1:M}$.
\subsection{Quantile Forecasting with Shared Linear Layers}
The time series forecasting is done with a Linear layer along the temporal axis. We follow the same base forecasting layer from \cite{zeng2022transformers}. This shall allow for a direct comparison
in terms of number of parameters and model capacity. More formally, for the QDLinear the forecasting can be expressed as: 
\begin{equation}
\begin{aligned}
    \hat{Y}_{i,trend} &= W_{trend} Z_{i,trend} + b_{trend} \\
    \hat{Y}_{i,season} &= W_{season} Z_{i,season} + b_{season} \\ 
    \hat{Y}_{i,QDLinear} &= \hat{Y}_{i,trend} + \hat{Y}_{i,season} 
\end{aligned}   
\end{equation}
$\hat{Y}_{i,QDLinear}$ represents forecast for $[\tau:T]$, for quantile level $\alpha'_i$. $W_{season},W_{trend} \in \mathbb{R}^{\len(1:\tau) \times \len(\tau:T)}, \mathbb{R}^{(\tau//w) \times \len(\tau:T)}$, $b_{season}, b_{trend} \in \mathbb{R}^{\len(\tau:T)}$ represent the learnable Linear layer parameters for forecasting. By repeating the same operation with a different quantile level embedding, we can generate the quantile forecasts for multiple quantile levels, $\{\hat{Y}_{i,QDLinear}\}_{i=1:M}$. Similarly, we can generate the quantile forecasts for the QNLinear normalization based input, $\{\hat{Y}_{i,QNLinear}\}_{i=1:M}$: 
\begin{equation}
\begin{aligned}
    \hat{Y}_{i,norm} &= W_{norm} Z_{i,norm} + b_{norm} \\
    \hat{Y}_{i,QNLinear} &= \hat{Y}_{i,norm}+ X_{\tau} 
\end{aligned}   
\end{equation}
Where, $W_{norm} \in \mathbb{R}^{\len(1:\tau) \times \len(\tau:T)}, b_{norm} \in \mathbb{R}^{\len(\tau:T)}$ are learnable layer parameters. The forecasts $\{\hat{Y}_{i,QLinear}\}_{i=1:M}$ for QLinear can be expressed as:
\begin{equation}
\begin{aligned}
    \hat{Y}_{i,linear} &= W_{linear} Z_{i,linear} + b_{linear} \\
\end{aligned}   
\end{equation}
Where, $W_{linear}, b_{linear}$ denote learnable parameters of same dimensionality as $W_{norm},b_{norm}$. 
Notably, we share the Linear layer parameters. We do this for all three models. We call the same Linear layers each time with a different input which contain different quantile embeddings added, as in Eq. \ref{eq:qdlinear}, \ref{eq:qnlinear}, and generate the respective quantile forecasts. This parameter sharing can be easily differentiated with having separate Linear layers per quantile level each tasked to forecast a different quantile level, leading to a significant increase in parameters. 
\subsection{Auxiliary Quantile Reconstruction}
In addition to auxiliary quantile forecasting, we also reconstruct the given time series input's for multiple quantile levels. Reconstruction has been well-established for standard supervised learning as an auxiliary task, and in this paper, we propose reconstruction of quantile levels for the time series input. This adds additional auxiliary tasks besides the auxiliary quantile forecasting tasks. For the reconstruction, we increase the dimensionality of the Linear layer parameters, for QDLinear, $W_{season},W_{trend} \in \mathbb{R}^{\len(1:\tau) \times \len(1:T)}$, $\mathbb{R}^{(\tau//w) \times \len(1:T)}$, $b_{season}, b_{trend} \in \mathbb{R}^{\len(1:T)}$; for QNLinear and QLinear, $W_{norm},W_{linear} \in \mathbb{R}^{\len(1:\tau) \times \len(1:T)}$ and $b_{norm},b_{linear} \in \mathbb{R}^{\len(1:T)}$. Note that, the auxiliary reconstruction also does not significantly increase parameters due to shared layers, however, there is still an increase in number of parameters for the layers based on the input length.
\subsection{Multi-task Loss Function}
For proposed models, the multi-task loss with quantile levels, 
\begin{align}
    \argmin_{\Theta \in \mathbb{R}} \sum_{n=1}^{N} \sum_{c=1}^{C} \sum_{t=1}^{T} \nonumber \rho_{\alpha_i=0.5}(Y^{n,c}_t, \hat{Y}^{n,c}_{t,\alpha_i=0.5}) \\ + \frac{1}{2(M-1)} \sum_{i=1}^{M-1} { \rho_{\alpha_i} (Y^{n,c}_t, \hat{Y}^{n,c}_{t,\alpha_i})) }
     \label{eq:mtl_loss_final}
\end{align}
Where, $\Theta$ denotes all learnable weights and $\rho_{\alpha} (Y, \hat{Y}_{\alpha_i})$ denotes the quantile loss, defined as:
\begin{align}
        \nonumber
        \rho_{\alpha_{i}}(Y, \hat{Y}_{\alpha_i}) &= (Y- \hat{Y}_{\alpha_i}) (\alpha_i - \mathbb{I}_{( Y\le \hat{Y}_{\alpha_i} )}) \\
                                                   &= \begin{cases}
                                                   \alpha_i(Y- \hat{Y}_{\alpha_i}), & \text{if } Y\geq  \hat{Y}_{\alpha_i},\\
                                                   (\alpha_i - 1) (Y- \hat{Y}_{\alpha_i}), &  \text{if } Y <  \hat{Y}_{\alpha_i},
                                                   \end{cases}
        \label{quantileloss}
\end{align}
Intuitively, the quantile loss is related to the $L_{1}$ loss, also refered to as the Least Absolute Deviation. Learning on the $L_1$ loss corresponds to learning the quantile level, $\alpha_{i}=0.5$ of the target distribution. Therefore, the quantile loss generalizes the $L_1$ loss to other quantile levels with specific parameterizations defined through quantile levels $\alpha_{1:M}$.

Importantly, multi-task loss functions are weighted, and these weights are tuned as hyperparameters. 
Therefore, we weight the auxiliary tasks less with the factor, $\frac{1}{2((M-1))}$.
For completeness, the loss in Eq. \ref{eq:mtl_loss_final} is normalized with $\frac{1}{N \times C \times T \times2}$. 

\begin{table*}[t]
\centering
\begin{adjustbox}{width=\textwidth}
\begin{tabular}{|c|c|c|ccc|ccccccccc|}
\hline
\multicolumn{1}{|l|}{}        & \multicolumn{1}{l}{} & \multicolumn{1}{l|}{} & \multicolumn{3}{c|}{Proposed}                    & \multicolumn{9}{c|}{Reported from \cite{zeng2022transformers}  }                                                                             \\ \hline
\multicolumn{1}{|c|}{Data} & Horizons             & Imp(\%)           & QDLin       & QNLin      & QLin      & DLin        & NLin        & Lin & FEDFor & Autofor & Infor & Pyrafor & LogTrans & Repeat         \\ \hline
\multirow{4}{*}{\rotatebox{90}{Electricity}}  & 96                   & 1.265                 & {\ul 0.235}    & {\ul 0.235}    & \textbf{0.234} & 0.237          & 0.237          & 0.237  & 0.308     & 0.317      & 0.368    & 0.449      & 0.357    & 0.946          \\
                              & 192                  & 0.806                 & \textbf{0.246} & \textbf{0.246} & {\ul 0.248}    & 0.249          & {\ul 0.248}    & 0.250  & 0.315     & 0.334      & 0.386    & 0.443      & 0.368    & 0.950          \\
                              & 336                  & 1.132                 & 0.264          & {\ul 0.263}    & \textbf{0.262} & 0.267          & 0.265          & 0.268  & 0.329     & 0.338      & 0.394    & 0.443      & 0.380    & 0.961          \\
                              & 720                  & 1.346                 & {\ul 0.294}    & 0.295          & \textbf{0.293} & 0.301          & 0.297          & 0.301  & 0.355     & 0.361      & 0.439    & 0.445      & 0.376    & 0.975          \\ \hline
\multirow{4}{*}{\rotatebox{90}{Exchange}}     & 96                   & -3.061                & 0.207          & {\ul 0.202}    & 0.204          & 0.203          & 0.208          & 0.207  & 0.278     & 0.323      & 0.752    & 1.105      & 0.812    & \textbf{0.196} \\
                              & 192                  & 1.038                 & \textbf{0.286} & 0.300          & {\ul 0.287}    & 0.293          & 0.300          & 0.304  & 0.380     & 0.369      & 0.895    & 1.151      & 0.851    & 0.289          \\
                              & 336                  & 1.515                 & 0.408          & 0.427          & \textbf{0.390} & 0.414          & 0.415          & 0.432  & 0.500     & 0.524      & 1.036    & 1.172      & 1.081    & {\ul 0.396}    \\
                              & 720                  & -18.136               & 0.813          & 0.710          & 0.795          & \textbf{0.601} & 0.780          & 0.750  & 0.841     & 0.941      & 1.310    & 1.206      & 1.127    & {\ul 0.681}    \\ \hline
\multirow{4}{*}{\rotatebox{90}{Traffic}}      & 96                   & 3.584                 & {\ul 0.270}    & 0.273          & \textbf{0.269} & 0.282          & 0.279          & 0.282  & 0.366     & 0.388      & 0.391    & 0.468      & 0.384    & 1.079          \\
                              & 192                  & 3.169                 & {\ul 0.276}    & \textbf{0.275} & {\ul 0.276}    & 0.287          & 0.284          & 0.287  & 0.373     & 0.382      & 0.379    & 0.467      & 0.390    & 1.087          \\
                              & 336                  & 4.137                 & 0.285          & {\ul 0.283}    & \textbf{0.278} & 0.296          & 0.290          & 0.295  & 0.383     & 0.337      & 0.420    & 0.469      & 0.408    & 1.095          \\
                              & 720                  & 4.234                 & {\ul 0.296}    & 0.300          & \textbf{0.294} & 0.315          & 0.307          & 0.315  & 0.382     & 0.408      & 0.472    & 0.473      & 0.396    & 1.097          \\ \hline
\multirow{4}{*}{\rotatebox{90}{Weather}}      & 96                   & 6.896                 & 0.261          & \textbf{0.216} & {\ul 0.217}    & 0.237          & 0.232          & 0.236  & 0.296     & 0.336      & 0.384    & 0.556      & 0.490    & 0.254          \\
                              & 192                  & 2.973                 & 0.324          & \textbf{0.261} & 0.280          & 0.282          & {\ul 0.269}    & 0.276  & 0.336     & 0.367      & 0.544    & 0.624      & 0.589    & 0.292          \\
                              & 336                  & 2.657                 & 0.316          & \textbf{0.293} & {\ul 0.294}    & 0.319          & 0.301          & 0.312  & 0.380     & 0.395      & 0.523    & 0.753      & 0.652    & 0.338          \\
                              & 720                  & 1.436                 & 0.375          & \textbf{0.343} & 0.351          & 0.362          & {\ul 0.348}    & 0.365  & 0.428     & 0.428      & 0.741    & 0.934      & 0.675    & 0.394          \\ \hline
\multirow{4}{*}{\rotatebox{90}{ILI}}          & 24                   & 1.981                 & 0.956          & \textbf{0.841} & 0.892          & 1.081          & {\ul 0.858}    & 0.985  & 1.260     & 1.287      & 1.677    & 2.012      & 1.444    & 1.701          \\
                              & 36                  & 5.355                 & 0.949          & \textbf{0.813} & 0.914          & 0.963          & {\ul 0.859}    & 1.036  & 1.080     & 1.148      & 1.467    & 2.031      & 1.467    & 1.884          \\
                              & 48                  & 0.904                 & 0.986          & \textbf{0.876} & 0.982          & 1.024          & {\ul 0.884}    & 1.060  & 1.078     & 1.085      & 1.469    & 2.057      & 1.468    & 1.798          \\
                              & 60                  & 4.580                 & 0.990          & \textbf{0.875} & 1.507          & 1.096          & {\ul 0.917}    & 1.104  & 1.157     & 1.125      & 1.564    & 2.100      & 1.560    & 1.677          \\ \hline
\multirow{4}{*}{\rotatebox{90}{ETTh1}}        & 96                   & 1.015                 & \textbf{0.390} & {\ul 0.394}    & 0.412          & 0.399          & {\ul 0.394}    & 0.397  & 0.419     & 0.459      & 0.713    & 0.612      & 0.740    & 0.713          \\
                              & 192                  & 0.481                 & 0.417          & \textbf{0.413} & 0.420          & 0.416          & {\ul 0.415}    & 0.429  & 0.448     & 0.482      & 0.792    & 0.681      & 0.824    & 0.733          \\
                              & 336                  & -1.639                & {\ul 0.434}    & 0.436          & 0.456          & 0.443          & \textbf{0.427} & 0.476  & 0.465     & 0.496      & 0.809    & 0.738      & 0.932    & 0.744          \\
                              & 720                  & 0.662                 & 0.482          & \textbf{0.450} & 0.520          & 0.490          & {\ul 0.453}    & 0.592  & 0.507     & 0.512      & 0.865    & 0.782      & 0.852    & 0.756          \\ \hline
\multirow{4}{*}{\rotatebox{90}{ETTh2}}       & 96                   & 1.775                 & {\ul 0.333}    & \textbf{0.332} & \textbf{0.332} & 0.353          & 0.338          & 0.352  & 0.388     & 0.397      & 1.525    & 0.597      & 1.197    & 0.422          \\
                              & 192                  & 0.787                 & {\ul 0.380}    & \textbf{0.378} & 0.385          & 0.418          & 0.381          & 0.413  & 0.439     & 0.452      & 1.931    & 0.683      & 1.635    & 0.473          \\
                              & 336                  & -0.499                & 0.415          & {\ul 0.402}    & 0.410          & 0.465          & \textbf{0.400} & 0.461  & 0.487     & 0.486      & 1.835    & 0.747      & 1.604    & 0.508          \\
                              & 720                  & 1.376                 & 0.453          & \textbf{0.430} & 0.463          & 0.551          & {\ul 0.436}    & 0.595  & 0.474     & 0.511      & 1.625    & 0.783      & 1.540    & 0.517          \\ \hline
\multirow{4}{*}{\rotatebox{90}{ETTm1} }       & 96                   & 2.915                 & \textbf{0.333} & {\ul 0.340}    & 0.350          & 0.343          & 0.348          & 0.352  & 0.419     & 0.475      & 0.571    & 0.510      & 0.546    & 0.665          \\
                              & 192                  & 1.369                 & \textbf{0.360} & {\ul 0.363}    & 0.370          & 0.365          & 0.375          & 0.369  & 0.441     & 0.496      & 0.669    & 0.537      & 0.700    & 0.690          \\
                              & 336                  & 0.777                 & {\ul 0.384}    & \textbf{0.383} & 0.386          & 0.386          & 0.388          & 0.393  & 0.459     & 0.537      & 0.871    & 0.655      & 0.832    & 0.707          \\
                              & 720                  & 0.950                 & 0.434          & \textbf{0.417} & \textbf{0.417} & {\ul 0.421}    & 0.422          & 0.435  & 0.490     & 0.561      & 0.823    & 0.724      & 0.820    & 0.729          \\ \hline
\multirow{4}{*}{\rotatebox{90}{ETTm2} }       & 96                   & 2.745                 & \textbf{0.248} & \textbf{0.248} & {\ul 0.250}    & 0.260          & 0.255          & 0.262  & 0.287     & 0.339      & 0.453    & 0.507      & 0.642    & 0.328          \\
                              & 192                  & 1.706                 & 0.295          & \textbf{0.288} & {\ul 0.289}    & 0.303          & 0.293          & 0.308  & 0.328     & 0.340      & 0.563    & 0.673      & 0.757    & 0.371          \\
                              & 336                  & 1.223                 & {\ul 0.324}    & \textbf{0.323} & 0.326          & 0.342          & 0.327          & 0.373  & 0.366     & 0.372      & 0.887    & 0.845      & 0.872    & 0.410          \\
                              & 720                  & 1.041                 & 0.386          & \textbf{0.380} & 0.385          & 0.421          & {\ul 0.384}    & 0.435  & 0.415     & 0.432      & 1.338    & 1.451      & 1.328    & 0.465          \\ \hline
Avg.                          & --                   & 1.236                 & 0.420          & \textbf{0.397} & 0.429          & 0.430          & {\ul 0.406}    & 0.443  & 0.490     & 0.515      & 0.886    & 0.858      & 0.864    & 0.759          \\ \hline
\# Wins                       & --                   & --                    & 6              & \textbf{21}    & {\ul 9}        & 1              & 2              & 0      & 0         & 0          & 0        & 0          & 0        & 1              \\ \hline
\end{tabular}
\end{adjustbox}
\caption{We report the performance comparison in terms of metric MAE. Best results are bold-faced, second-best underlined. The Improvement compares the best proposed result to the best per dataset and prediction length task from the benchmark in \cite{zeng2022transformers}}
\label{tab:results}
\end{table*}

\section{Experiments\protect\footnote{ \texorpdfstring{\MakeLowercase{github.com/super-shayan/qlinear}}{github.com/super-shayan/qlinear}    }}
We provide extensive experimental results on well-established datasets and forecasting tasks. A brief summary of the dataset statistics is given in Table \ref{tab:datasets}. 
Notably, for each of the proposed methods, we tune the number of auxiliary forecasting tasks per forecast horizon, $M$, as a hyperparameter in the grid, \{1,2, 4, 6, 8, 16, 32, 64, 128, 256, 512, 1024\}. Intuitively, forecasting a single quantile level $M=1$, implicitly, is similar to the Implicit quantile learning method from \cite{dabney2018implicit,gouttes2021probabilistic}. We schedule experiments for each dataset and prediction length task, with the number of auxiliary tasks as a hyperparameter covering the values in the grid noted above. The final reported test performance corresponds to the least validation split\footnote{Train, validation and test splits are derived chronologically splitting the time series in ratio 6:2:2 for ETT datasets, and 7:1:2 for the other datasets.} error based on the number of auxiliary quantiles' hyperparameter, $M$. Note that we query the forecasts at $\alpha_i=0.5$ quantile level for dataset test splits. Given that the proposed methods, QDLinear, QNLinear and QLinear correspond to the baseline architectures DLinear\cite{zeng2022transformers}, NLinear\cite{zeng2022transformers} and Linear\cite{zeng2022transformers} respectively, we keep all the rest of the hyperparameters and seeds same between proposed methods and their counterparts \cite{zeng2022transformers}. Hence, this allows us to examine the advantage of auxiliary quantile forecasting and reconstruction. Nevertheless, in cases where the number of quantiles greatly increase, intuitively a deeper encoded representation can lead to further performance gains, which is currently fixed across different quantile levels. \\ \indent The results are stated in Table. \ref{tab:results} and include the state-of-the-art baselines, FEDFormer \cite{zhou2022fedformer}, Autoformer\cite{wu2021autoformer}, Informer\cite{zhou2021informer}, Pyraformer\cite{liu2021pyraformer} and LogTrans\cite{li2019enhancing}. A non-learned Repeat prediction heuristic\cite{zeng2022transformers} is also compared against. We can observe that the proposed methods are able to outperform the baselines on most datasets and come to a close second otherwise. The performance gains are also consistent across the different prediction lengths, with significant improvement percentages averaging 1.236\%. Besides, we observed that performances varied across the different number of quantile levels for all datasets and forecast horizons, validating the effectiveness of tuning it as a hyperparameter.
\section{Conclusion}
Our multi-task learning methodology to forecast auxiliary quantiles lead to superior results. It also does not significantly increase the learning parameters compared to the direct counterpart Linear architectures via utilizing shared parameters. Future work includes applications to univariate data and researching non-linear modeling of quantile levels and input.
\section{Acknowledgements}
This work is co-funded by the industry project Data-driven Mobility Services of ISMLL and Volkswagen Financial Services; also through BMWK Germany project, IIP-Ecosphere: Next Level Ecosphere for Intelligent Industrial Production.
\bibliographystyle{IEEEbib}
\bibliography{strings}

\begin{thebibliography}{10}

\bibitem{zeng2022transformers}
Ailing Zeng, Muxi Chen, Lei Zhang, and Qiang Xu,
\newblock ``Are transformers effective for time series forecasting?,''
\newblock {\em arXiv preprint arXiv:2205.13504}, 2022.

\bibitem{wu2021autoformer}
Haixu Wu, Jiehui Xu, Jianmin Wang, and Mingsheng Long,
\newblock ``Autoformer: Decomposition transformers with auto-correlation for
  long-term series forecasting,''
\newblock {\em Advances in Neural Information Processing Systems}, vol. 34, pp.
  22419--22430, 2021.

\bibitem{zhou2021informer}
Haoyi Zhou, Shanghang Zhang, Jieqi Peng, Shuai Zhang, Jianxin Li, Hui Xiong,
  and Wancai Zhang,
\newblock ``Informer: Beyond efficient transformer for long sequence
  time-series forecasting,''
\newblock in {\em Proceedings of AAAI}, 2021.

\bibitem{koenker2001quantile}
Roger Koenker and Kevin~F Hallock,
\newblock ``Quantile regression,''
\newblock {\em Journal of economic perspectives}, vol. 15, no. 4, pp. 143--156,
  2001.

\bibitem{wen2017multi}
Ruofeng Wen, Kari Torkkola, Balakrishnan Narayanaswamy, and Dhruv Madeka,
\newblock ``A multi-horizon quantile recurrent forecaster,''
\newblock {\em arXiv preprint arXiv:1711.11053}, 2017.

\bibitem{park2021learning}
Youngsuk Park, Danielle Maddix, Fran{\c{c}}ois-Xavier Aubet, Kelvin Kan, Jan
  Gasthaus, and Yuyang Wang,
\newblock ``Learning quantile functions without quantile crossing for
  distribution-free time series forecasting,''
\newblock {\em arXiv preprint arXiv:2111.06581}, 2021.

\bibitem{dabney2018implicit}
Will Dabney, Georg Ostrovski, David Silver, and R{\'e}mi Munos,
\newblock ``Implicit quantile networks for distributional reinforcement
  learning,''
\newblock in {\em International conference on machine learning}. PMLR, 2018,
  pp. 1096--1105.

\bibitem{gouttes2021probabilistic}
Adele Gouttes, Kashif Rasul, Mateusz Koren, Johannes Stephan, and Tofigh
  Naghibi,
\newblock ``Probabilistic time series forecasting with implicit quantile
  networks,''
\newblock {\em arXiv}, 2021.

\bibitem{zhou2022fedformer}
Tian Zhou, Ziqing Ma, Qingsong Wen, Xue Wang, Liang Sun, and Rong Jin,
\newblock ``Fedformer: Frequency enhanced decomposed transformer for long-term
  series forecasting,''
\newblock {\em arXiv preprint arXiv:2201.12740}, 2022.

\bibitem{liu2021pyraformer}
Shizhan Liu, Hang Yu, Cong Liao, Jianguo Li, Weiyao Lin, Alex~X Liu, and
  Schahram Dustdar,
\newblock ``Pyraformer: Low-complexity pyramidal attention for long-range time
  series modeling and forecasting,''
\newblock in {\em International Conference on Learning Representations}, 2021.

\bibitem{li2019enhancing}
Shiyang Li, Xiaoyong Jin, Yao Xuan, Xiyou Zhou, Wenhu Chen, Yu-Xiang Wang, and
  Xifeng Yan,
\newblock ``Enhancing the locality and breaking the memory bottleneck of
  transformer on time series forecasting,''
\newblock {\em Advances in Neural Information Processing Systems (NeurIPS)},
  2019.

\end{thebibliography}
\end{document}